\def\year{2021}\relax
\documentclass[letterpaper]{article} 
\usepackage{aaai21}  
\usepackage{times}  
\usepackage{helvet} 
\usepackage{courier}  
\usepackage[hyphens]{url}  
\usepackage{graphicx} 
\usepackage{natbib}  
\usepackage{caption} 
\usepackage{graphicx}
\usepackage{amsmath,amssymb} 
\usepackage{color}
\usepackage{multirow}
\usepackage[dvipsnames]{xcolor}
\usepackage[switch]{lineno}

\usepackage{colortbl}

\definecolor{lightskyblue}{rgb}{0.94,1.0,1.0}
\definecolor{lightgreen}{rgb}{0.94,1.0,0.94}
\definecolor{whitesmoke}{rgb}{0.92,0.92,0.92}
\definecolor{seashell}{rgb}{1.0,0.96,0.93}

\begin{document}
\title{Exploiting Diverse Characteristics and Adversarial Ambivalence for Domain Adaptive Segmentation}

\author{
Bowen Cai\textsuperscript{1}\thanks{Corresponding author.} \ \ Huan Fu\textsuperscript{1} \ \ Rongfei Jia\textsuperscript{1} \ \ Binqiang Zhao\textsuperscript{1} \ \ Hua Li\textsuperscript{2} \ \ Yinghui Xu\textsuperscript{1} 
\vspace{1.8mm}
\\
{\normalfont\textsuperscript{1}Alibaba Group} \\
{\normalfont\textsuperscript{2}Institute of Computing Technology, Chinese Academy of Sciences}\\
{\tt\small \{kevin.cbw, fuhuan.fh, rongfei.jrf, binqiang.zhao\}@alibaba-inc.com} \\
{\tt\small \{lihua\}@ict.ac.cn}
}

\maketitle

\begin{abstract}

Adapting semantic segmentation models to new domains is an important but challenging problem. Recently enlightening progress has been made, but the performance of existing methods are unsatisfactory on real datasets where the new target domain comprises of heterogeneous sub-domains (\emph{e.g.,} diverse weather characteristics). We point out that carefully reasoning about the multiple modalities in the target domain 
can improve the robustness of adaptation models. To this end, we propose a condition-guided adaptation framework that is empowered by a special attentive progressive adversarial training (APAT) mechanism and a novel self-training policy. The APAT strategy progressively performs condition-specific alignment and attentive global feature matching. The new self-training scheme exploits the adversarial ambivalences of easy and hard adaptation regions and the correlations among target sub-domains effectively. We evaluate our method (DCAA) on various adaptation scenarios where the target images vary in weather conditions. The comparisons against baselines and the state-of-the-art approaches demonstrate the superiority of DCAA over the competitors.
\end{abstract}

\section{Introduction}
\label{sec:intro}

\noindent  Due to the expensive labeling costs, there is a growing interest in studying weakly-supervised or unsupervised approaches for semantic segmentation in recent years. One of the widely investigated techniques is synthetic-to-real unsupervised domain adaptation (UDA) \cite{hoffman2016fcns,zhang2017curriculum,tsai2018learning,zou2019confidence,lambert2020mseg}, which expects to transfer knowledge from labeled synthetic datasets to unlabeled real domains. Recent solutions can be mainly divided into two streams, including feature alignment and self-training. The former utilizes adversarial training to align both the image styles and semantic feature distributions. As a promising alternative, self-training focuses on iteratively generating pseudo-labels with high confidence and fewer noises for target images, and retrain the networks using the produced labels. However, the performance of these models is still far from satisfactory on real datasets. One possible reason is that existing methods assume that images in the target domain are captured under similar conditions, while a new real domain can have diverse characteristics. For example, the images taken in the same place with the same viewpoint may show significant differences because of weather changes. 


The above analysis motivates us to study a more realistic UDA setting for semantic segmentation, i.e.,  the images in a target domain has diverse characteristics. Taking ``weather" as an example, a naive idea is to enlarge the synthetic dataset to cover as many weather conditions as possible and then directly adopt existing high-performing UDA methods. Unluckily, we find that the performance gains are not that remarkable. A possible reason is that the adversarial adaptation learning process may be more challenging when images in both domains show multiple modalities, resulting in semantic distortion translations and ambiguous feature alignment near weather condition boundaries. In fact, minimizing the discrepancy between synthetic and real domains is challenging \cite{saito2018maximum}, not to mention that each domain consists of several sub-domains (\emph{e.g.}, Cityscapes$^{*}$: $\{Cloudy, Foggy, Rainy\}$). Moreover, the learned pseudo-labels for target images contain significant noise when the atmosphere states are unconstrained.


\begin{figure*}[th!]
\centering
\includegraphics[scale=0.43]{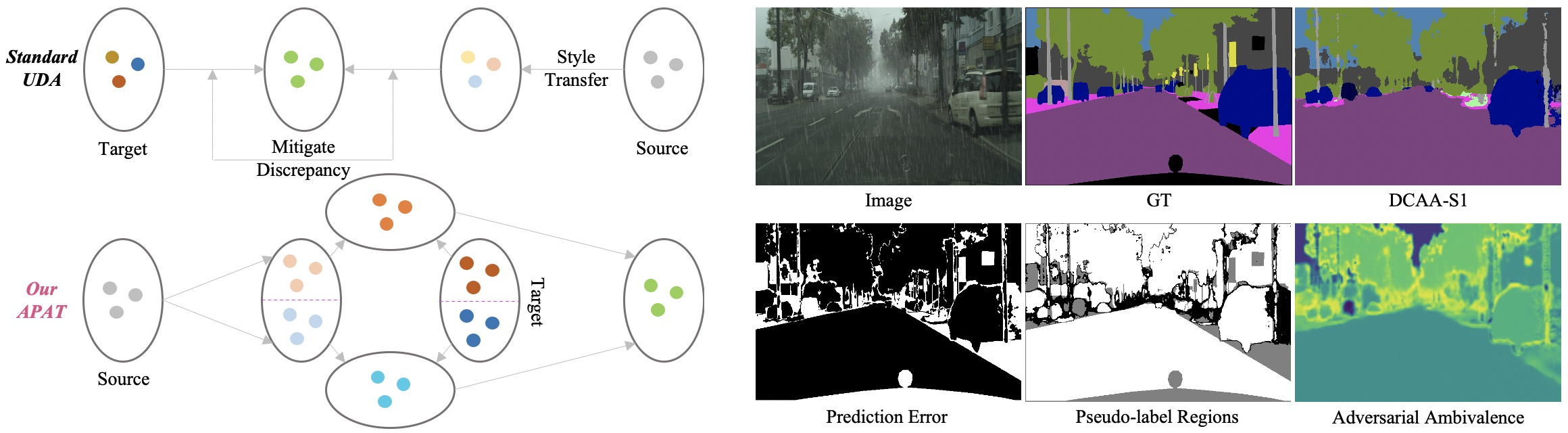}
\caption{\textbf{Left:} A comparison between the typical domain adaptation process (Left Top) and our attentive progressive adversarial training (APAT) scheme (Left Bottom). \textbf{Right:} Adversarial Ambivalence and Pseudo-label. DCAA-S1: The first-stage prediction of DCAA. Prediction Error: The inconsistency between GT and DCAA-S1 (White). Pseudo-label Regions: ``Black" represents the hard adaptation regions (without pseudo-labels); ``Gray" and ``White" denote regions with incorrect and correct pseudo-labels, respectively. Adversarial Ambivalence: The adversarial confidence obtained from the discriminator. ``Yellow" means the regions are not well adapted. We find that the adversarial ambivalence shows consistency with the prediction error, thus can revise the pseudo-label related constraints. We refer to the ``Introduction" section for more analyses of our motivations.}
\label{fig:motivation}
\end{figure*}

In this paper, we argue that carefully reasoning about the diverse characteristics in the target domain may improve the effectiveness of adapted segmentation model. Towards this goal, we propose a condition guided adaptation framework (DCAA) empowered by a special attentive progressive adversarial training (APAT) mechanism and a novel self-training policy. We illustrate our observations and idea in Figure~\ref{fig:motivation}. For the APAT strategy, we first introduce a condition-guided style translator to generate realistic characteristic conditionable samples from synthetic images. Furthermore, we develop a condition attention module to exploit the complementarities of multi-modalities. Lastly, we study condition-specific adversarial training to align semantic feature distributions under each condition and further mitigate the global discrepancy. 

For the novel self-training scheme, we investigate the correlations among target sub-domains to produce robust attentive pseudo-labels for target images. Importantly, we point out that the dense output from the discriminators implies adversarial ambivalence for hard adaptation regions. We thus consider the ambivalence to further improve the self-training process by revising the pseudo-labels and emerging the hard adaptation regions. We evaluate our DCAA on several challenge adaptation scenarios, including  GTA5$\to$Cityscapes$^{*}$ \citep{Richter_2016_ECCV,Cordts2016Cityscapes,sakaridis2018model,hu2019depth}, SYNTHIA$\to$Cityscapes$^*$ \citep{Ros_2016_CVPR}, and GTA5$\to$BDD100K \citep{yu2020bdd100k}. Qualitative and quantitative comparisons against baselines and SOTA UDA methods demonstrate the superiority of our DCAA.

\section{Related Work}

\noindent Recent unsupervised domain adaptive segmentation solutions mainly focus on two points, \emph{i.e.}, feature alignment \cite{du2019ssf,yang2020fda,wang2020differential,kim2020learning,xiao2020multi,licontent,yang2019adversarial,zhang2020joint,chen2019joint,wang2018deep} and self-training \cite{zhou2017scene,zhang2017curriculum,pan2020unsupervised,pei2018multi,chen2019domain,lian2019constructing,vu2019advent}. Among the feature alignment approaches, FCN in the wild \cite{hoffman2016fcns} was the first to introduce adversarial learning for global feature alignment and category statistic matching. AdaptSegNet \cite{tsai2018learning} modeled the structured spatial similarities on the semantic output space. Subsequently, many promising strategies has been devised to learn domain invariant features, such as conditional generator \cite{hong2018conditional}, task-specific decision boundaries \cite{saito2018maximum}, co-training \cite{luo2019taking}, and patch distribution matching \cite{tsai2019domain}. Other works show that style transfer \cite{zhu2017unpaired} could largely mitigate the domain gap \cite{zhang2018fully,sun2019not}. 
Thus, researchers exploited techniques such as semantic consistency \cite{hoffman2017cycada}, channel-wise feature alignment \cite{wu2018dcan}, symmetric adaptation consistency \cite{chen2019crdoco}, and controllable translation degrees \cite{gong2019dlow} to improve both the style translation and adaption processes. Self-training has recently gained considerable interests for UDA. For example, CBST \cite{zou2018unsupervised} introduced an iterative class-balanced self-placed learning policy to alternatively performed balanced pseudo-label generation and model retraining. The following works demonstrated the superiority of self-training by studying confidence regularization \cite{zou2019confidence}, bidirectional adversarial translation and self-supervised learning \cite{li2019bidirectional}, and category-wise centroids \cite{zhang2019category}. These works always overlooked the diverse characteristics of natural images in a target domain, thus might produce sensitive models. 

There are two adaptation settings that are closely related to our studied setting, \emph{i.e.}, multi-target domain adaptation (MTDA) \cite{gholami2020unsupervised} and compound domain adaptation (CDA) \cite{liu2020open,chen2019blending}.  In MTDA, researchers expected to learn a general domain-agnostic model for largely unconstrained target domains. For example, the widely studied PACS dataset \cite{li2017deeper} contains domains such as photo, cartoon, sketch, and art painting. In CDA, algorithms were developed to discover the unknown sub-domains in a target domain. In contrast, we study the setting where the sub-domains (from a single target domain) are strongly correlated in content, and the sub-domain labels are available (\emph{e.g.}, BDD100K). We mainly take efforts into modeling the correlations (complementarities) of the multi-modalities. Thus, our proposed approach is largely different from these MTDA and CDA methods. We clarify that assigning image-level (sub-domain) labels to a target domain is inexpensive in practice. We could even learn a condition classification network for the purpose.

\begin{figure*}[th!]
\centering
\includegraphics[scale=0.4]{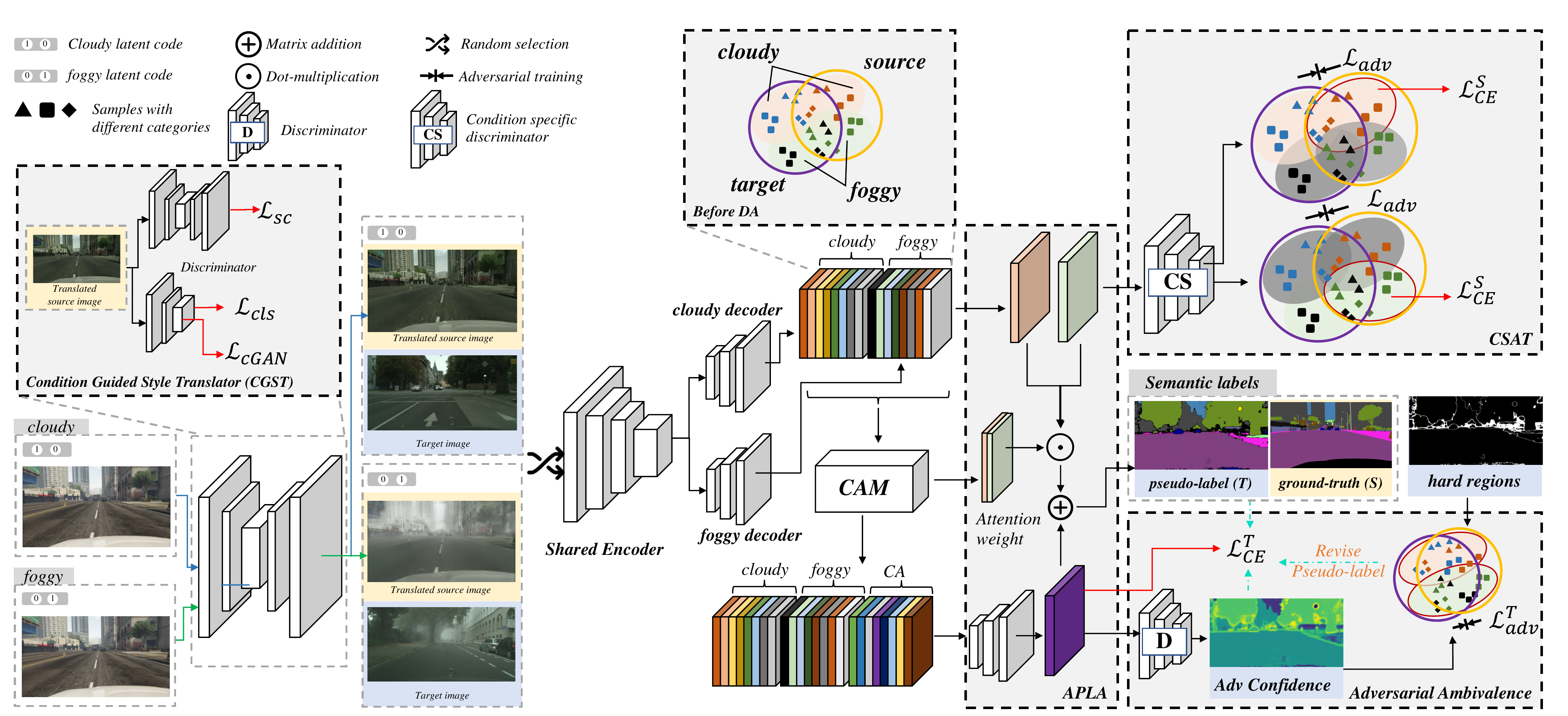}
\caption{\textbf{DCAA.} We take GTA5$\to$Cityscapes: $\{Cloudy, Foggy\}$ as an example to illustrate our DCAA. The framework is empowered by a special attentive progressive adversarial training scheme (CGST + CAM + CSAT) and a novel self-training policy (APLA + Adversarial Ambivalence).}
\label{fig:network}
\end{figure*}

\section{Methodology}
\noindent In the section, we take a special case that the target images are under two weather conditions to illustrate our method. Let $X^S = \{x^s_i\}_{i=1}^N$ and $Y^S = \{y^s_i\}_{i=1}^N$ be the source images and corresponding dense labels. The target images consist of two parts, \emph{i.e.}, $X^T = \{X^{T1}, X^{T2}\}$, where $T_1$ and $T_2$ represent the special environmental conditions in this case. Our goal is to transfer knowledge from $S$ to $T$ so that a learned segmentation model can produce promising parsing for images in the target domain. To this end, we develop a condition guided adaptation framework (DCAA), as shown in Figure~\ref{fig:network}. In the following, we will present our attentive progressive adversarial training (APAT) mechanism and self-training policy in detail.

\subsection{Attentive Progressive Adversarial Training}
\noindent Previous works \cite{luo2019taking,hoffman2017cycada,tsai2018learning,tsai2019domain} preferred to take adversarial training to straightforwardly mitigate the source and target discrepancy. In contrast, we find that progressively reducing the domain shift by considering the weather conditions in the target domain would improve the adaptation process. Here, our APAT solution are constructed with three policies including condition guided style transfer, condition attention, and condition-specific adversarial training.

\subsubsection{Condition Guided Style Transfer (CGST).} We are here to learn a conditional generator $G_{ST}(X^S, c): X^S \to X^T = \{X^{T_1}, X^{T_2}\}$ such that $G_{ST}(X^S)$ has the same distribution as $X^{T_1}$ or $X^{T_2}$ according to the specific latent variable $c \in \{0, 1\}$ (represents $\{T_1, T_2\}$). Thus, we develop a condition guided style transfer (CGST) following standard conditional GANs \cite{mirza2014conditional} coupled with a semantic consistency constraint. We borrow concepts in StarGAN \cite{choi2018stargan} and BicycleGAN \cite{zhu2017toward} to improve the translation. 


Specifically, we adopt a multi-level concatenation strategy \cite{zhu2017toward} to inject $c$ into $G_{ST}$ as the translation guidance. We perform adversarial training to enforce $G_{ST}$ producing realistic images with preferred styles. The adversarial constraint takes the form as:
\begin{equation}
    \begin{aligned}
      \mathcal{L}_{cGAN} &= \mathbb{E}_{x^t \sim X^{T}}[\log(D_T(x^t))] \\
      & + \mathbb{E}_{x^s \sim X^{S}}[\log(1 - D_T(G_{ST}(x^s, c)))],
    \end{aligned}
\end{equation}
where $D_T$ is the discriminator with respect to $X^T$. The guidance from the injecting operation alone may not guarantee weather conditionable translations. Thus, we add an auxiliary condition classification loss \cite{choi2018stargan} to further constraint the solution space:
\begin{equation}
    \begin{aligned}
      \mathcal{L}_{cls} &= \mathbb{E}_{x \sim X^T, c}[-\log(D_{cls}(c|x))] \\
      & + \mathbb{E}_{x \sim X^T, c}[-\log(D_{cls}(c|G_{ST}(x)))],
    \end{aligned}
\end{equation}
where $D_{cls}$ is the condition classifier which shares the same backbone with $D^{T}$. 

We find that the obtained translator driven by constraints above would generate images with semantic distortion or inconsistency. For example, the trees in source images may become unreasonable building materials after the translation. We provide a remedy by incorporating a semantic consistency constraint which can be expressed as:
\begin{equation}
    \begin{aligned}
      \mathcal{L}_{sc} &= \mathbb{E}_{(x,y) \sim (X^S, Y^S)}[-\log(F_{seg}(y|G_{ST}(x, c)))],
    \end{aligned}
\end{equation}
where $F_{seg}$ is a segmentation network. We use a light version of UperNet \cite{xiao2018unified} in the paper.

The full objective for our CGST can be expressed as:
\begin{equation}
    \begin{aligned}
      \mathcal{L}_{CGST} &= \mathcal{L}_{cGAN} + \mathcal{L}_{cls} + \lambda_{sc}\mathcal{L}_{sc},
    \end{aligned}
\end{equation}
where the trade-off parameter $\lambda_{sc}$ is set to $5.0$ in our paper.

\begin{figure}[th!]
\centering
\includegraphics[scale=0.36]{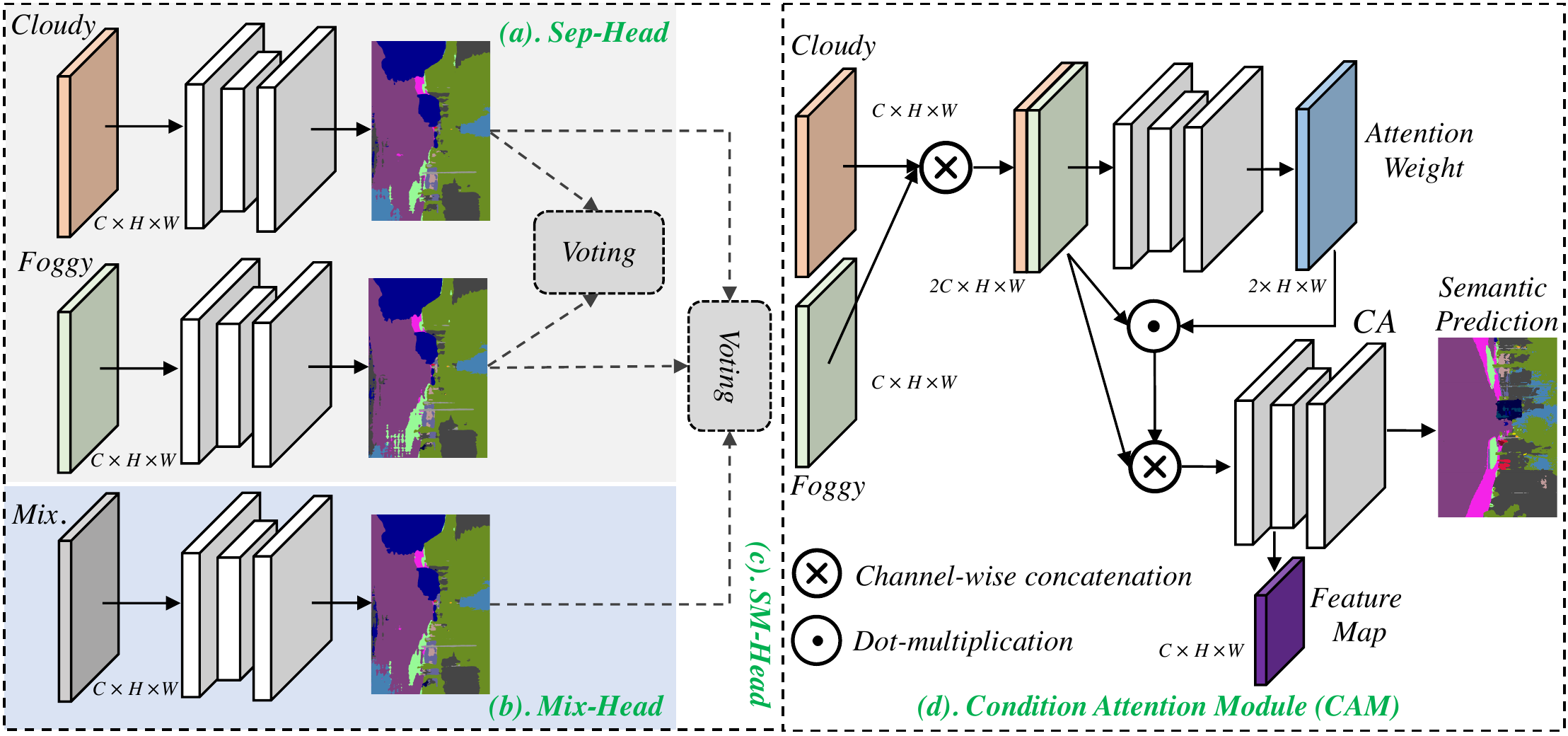}
\caption{Several options in incorporating weather condition priors, including Sep-Head, Mix-Head, SM-Head, and our CAM. Dot lines refer that we use mean voting to capture the final prediction in the inference stage.}
\label{fig:cam}
\end{figure}

\subsubsection{Condition Attention Module (CAM).} After obtaining the stylized images $\hat{X}^{S} = \{\hat{X}^{S_1}, \hat{X}^{S_2}\} = G_{ST}(X^{S}, c)$, the trivial solution is to learn some multi-head networks to explicitly incorporate weather priors, as shown in Figure~\ref{fig:cam}. We clarify that the effortless procedures may not sufficiently exploit the correlations of multi-modalities. We thus develop the condition attention module to address the issue. 

In detail, we feed $x \in \{\hat{X}^S, X^T\}$ into a shared encoder (\emph{Enc}) and two individual decoders (\emph{Dec}$^{C_1}$ and \emph{Dec}$^{C_2}$) to capture:
\begin{equation}
    \begin{aligned}
      f^{C_1}(x),\, p^{C_1}(x) &= Dec^{C_1}(Enc(x)), \\
      f^{C_2}(x),\, p^{C_2}(x) &= Dec^{C_2}(Enc(x)), \\
    \end{aligned}
\end{equation}
where $f^{C_1}(x),\, f^{C_2}(x) \in R^{D \times H \times W}$ are the semantic feature maps with dimension $D$, and $p^{C_1}(x),\, p^{C_2}(x) \in R^{L \times H \times W}$ are the probability predictions. In our paper, $D$ is set to 256, and $L=19$ is the number of semantic categories. Then, we learn attention weights $\mathcal{W} \in R^{2 \times H \times W}$ with a convolutional block (\emph{Conv}) from $f^{C_1}(x)$ and $f^{C_2}(x)$ to capture the enhanced feature $f$. In further, we concatenate $f^{C1}(x)$, $f^{C2}(x)$, and $f$ together, and feed them to an additional parsing head (\emph{CA}) to secure both $f^{CA}(x)$ and $p^{CA}(x)$. The operations can be expressed as:
\begin{equation}
    \begin{aligned}
      & \mathcal{W} = Conv(f^{C_1}(x) \oplus f^{C_2}(x)), \\
      & f = \mathcal{W}^{C_1} \otimes f^{C_1}(x) + \mathcal{W}^{C_2} \otimes f^{C_2}(x), \\
      & f^{CA}(x),\, p^{CA}(x) = CA(f^{C_1}(x) \oplus f^{C_2}(x) \oplus f),
    \end{aligned}
\end{equation}
where $\oplus$ is the concatenation operation, $\otimes$ is the element-wise product operation, and $\mathcal{W}^{Ci}$ is the $i$th channel of $\mathcal{W}$. 

\subsubsection{Condition-Specific Adversarial Training (CSAT).}
To optimize the network, we introduce condition-specific adversarial training to encourage the decoders, \emph{i.e.}, $Dec^{C_1}$, $Dec^{C_2}$, and $CA$, to focus on different characteristics. This allows alleviating the discrepancy between source and target feature distributions in a progressive manner. Specially, we employ three discriminators ($D^{C_1}$, $D^{C_2}$, and $D^{CA}$) for each decoder, respectively. $D^{C_i}$ addresses the specific condition $i$, thus operates on $\hat{X}^{S_i}$ and $X^{T_i}$ to match the distributions $p^{C_i}(\hat{X}^{S_i})$ and $p^{C_i}(X^{T_i})$. And $D^{CA}$ targets at transferring knowledge from $\hat{X}^S$ to $X^T$, thus is a standard domain classifier. The encoder and decoders are optimized by:
\begin{equation}
    \begin{aligned}
      \mathcal{L}_{adv}(x^t) &= - \sum_{h,w}\log(D^{CA}(p^{CA}(x_{hw}^t))) \\
      & - \sum_{i}\delta(x^t \in X^{T_i})\sum_{h,w}\log(D^{C_i}(p^{C_i}(x_{hw}^t))),
    \end{aligned}
\end{equation}
where $\delta(\cdot)$ is an indicator function which satisfies $\delta(True) = 1$ and $\delta(False) = 0$. Our discriminators follow the PatchGAN \cite{isola2017image} fashion. Note that, instead of learning source-to-target adaptation, we perform target-to-source adaptation so that we can obtain the adversarial ambivalences for target images. We will introduce the adversarial ambivalence later.

Importantly, with the provided source-domain $\hat{X}^S$ and $Y^S$ pairs, we have a corresponding condition-specific segmentation loss:
\begin{equation}
    \begin{aligned}
      \mathcal{L}^{S}_{CE}(\hat{x}^s) &= - \sum_{hw} \sum_{l=1}^L [y_{hwl}^s\log(p^{CA}_{hwl}(\hat{x}^s) + \\
      & \sum_{i}\delta(\hat{x}^s \in \hat{X}^{S_i}) y_{hwl}^s\log(p^{C_i}_{hwl}(\hat{x}^s))].
    \end{aligned}
\end{equation}

\subsection{Self-training}
\label{sec:self-training}
The proposed self-training policy aims to produce reliable pseudo-labels for target images and study adversarial ambivalence to revise the pseudo-labels and emerging the hard adaptation regions.

\subsubsection{Attentive Pseudo-Label Assignment (APLA).} 
Benefiting from the attentive progressive adversarial training scheme, we can obtain an adaptation model $M_{0}$. As a result, we can secure pseudo-labels for target images where the correctness is guaranteed to a certain degree. In detail, given a target image $x^t$, $M_{0}$ will produce predictions $p^{C_1}(x^t)$, $p^{C_2}(x^t)$, and $p^{CA}(x^t)$, as well as the attention weights $\mathcal{W}(x^t)$. 
The attentive pseudo-labels ${y}^t$ are generated as: 
\begin{equation}
    \begin{aligned}
      p(x^t) &= 0.5*\sum_{i=1}^K\mathcal{W}^{C_i}(x^t)\otimes p^{C_i}(x^t) + 0.5*p^{CA}(x^t) \\
      y_{hw}^t (x^t) &=
      \begin{cases}
      l & l = \mathop{\arg\max}_{l}\, p_{hw}(l|x^t) \\
      & \&\ \ p_{hw}(l|x^t) > \lambda_p \\
      0 & \text{otherwise}
      \end{cases},
    \end{aligned}
\end{equation}
where $hw$ is the spatial location, $l$ represents the category label, and $\lambda_p$ denotes the pre-defined threshold. 

\setlength\tabcolsep{3pt}
\begin{table*}[th!]
\footnotesize
\centering
\begin{tabular}{ c | c c c c c c c c c c c c c c c c c c c | c }
\hline
Method & \rotatebox{90}{Road} & \rotatebox{90}{SW} & \rotatebox{90}{Build} & \rotatebox{90}{Wall} & \rotatebox{90}{Fence} & \rotatebox{90}{Pole} & \rotatebox{90}{Light} & \rotatebox{90}{Sign} & \rotatebox{90}{Veg.} & \rotatebox{90}{Terrace} & \rotatebox{90}{Sky} & \rotatebox{90}{Person} & \rotatebox{90}{Rider} & \rotatebox{90}{Car} & \rotatebox{90}{Truck} & \rotatebox{90}{Bus} & \rotatebox{90}{Train} & \rotatebox{90}{Moter} & \rotatebox{90}{Bike} & \rotatebox{90}{\textcolor{red}{mIoU}} \\
\hline\hline
\multicolumn{21}{c}{GTA5 $\to$ Cityscapes$^{*}$} \\
\hline
Source Only & 50.1 & 17.1 & 51.7 & 19.7 & 17.6 & 20.4 & 22.5 & 2.6 & 76.6 & 18.1 & 72.1 & 56.0 & 27.1 & 70.8 & 4.7 & 8.7 & 0.2 & 11.1 & 35.1 & 30.6 \\
AdaptSegNet & 80.4 & 16.4 & 71.9 & 14.3 & 17.2 & 18.0 & 30.1 & 20.3 & 75.3 & 15.0 & 74.8 & 53.0 & 26.0 & 76.0 & 12.8 & 32.4 & 2.7 & 24.1 & 29.5 & 36.3 \\
CLAN &86.6 & 12.7 & 74.0 & 18.6 & 16.0 & 22.7 & 32.6 & 20.1 & 73.3 & 25.8 & 76.6 & 54.5 & 28.0 & 80.3 & 24.3 & 31.8 & 0.1 & 18.5 & 25.8 & 38.0 \\
AdvEnt & 88.2 & 33.1 & 78.3 & 26.0 & 11.1 & 31.3 & 34.7 & 26.3 & 74.5 & 22.0 & 63.4 & 55.3 & 29.9 & 82.6 & 25.0 & 25.0 & 0.01 & 20.0 & 31.7 & 39.9 \\
CRST  & \textbf{90.6} & 42.8 & 78.5 & 21.0 & 12.1 & 35.1 & 35.8 & 25.9 & 69.7 & 14.6 & 78.2 & 61.3 & 30.2 & 81.7 & 16.0 & 34.8 & 5.04 & \textbf{28.7} & 47.5 & 42.6  \\
CAG-UDA & 87.1 & 37.9 & 82.7 & 26.3 & 29.9 & 35.1 & 41.2 & 39.1 & 82.3 & 27.8 & 76.3 & 61.6 & 32.1 & 83.6 & 17.8 & 34.3 & 18.8 & 19.2 & 30.1 & 45.4 \\
AMEAN++ & 88.0 & 41.5 & 81.1 & 36.5 & 27.3 & 30.7 & 45.0 & 27.3 & 82.6 & 37.3 & 73.5 & 61.7 & 32.4 & 85.0 & 27.6 & 42.1 & 13.6 & 17.5 & 23.7 & 46.0 \\
OCDA++ & 73.5 & 29.8 & 67.7 & 18.7 & 22.7 & 26.7 & 35.9 & 22.4 & 81.1 & 32.3 & 72.2 & 48.1 & 25.6 & 78.1 & 11.2 & 12.8 & 1.8 & 20.7 & 25.8 & 37.2 \\
MDTA-ITA & 80.1 & 23.6 & 76.0 & 17.6 & 24.2 & 25.4 & 34.5 & 14.2 & 77.7 & 18.3 & 73.5 & 49.7 & 12.3 & 79.7 & 16.9 & 22.9 & 5.1 & 6.8 & 10.8 & 35.2 \\
\rowcolor{lightgreen}
\textcolor{Orange}{DCAA} & 90.5 & \textbf{52.8} & \textbf{84.9} & \textbf{38.7} & \textbf{30.4} & \textbf{37.6} & \textbf{45.5} & \textbf{42.4} & \textbf{84.5} & \textbf{45.5} & \textbf{80.7} & \textbf{63.8} & \textbf{36.2} & \textbf{86.3} & \textbf{34.9} & \textbf{45.1} & \textbf{19.7} & 14.9 & \textbf{42.5} & \textbf{51.4} \\
\hline\hline
\multicolumn{21}{c}{SYNTHIA $\to$ Cityscapes$^{*}$} \\
\hline
Source Only & 58.6 & 21.3 & 65.7 & 8.0 & 0.0 & 22.3 & 1.6 & 9.3 & 70.1 & - & 71.7 & 58.4 & 25.9 & 64.1 & - & 10.7 & - & 14.1 & 24.6 & 32.9 \\
AdaptSegNet & \textbf{84.7} & \textbf{44.1} & 74.1 & 8.7 & 0.2 & 17.5 & 3.8 & 4.3 & 71.3 & - & 69.5 & 47.8 & 16.4 & 62.2 & - & 23.5 & - & 10.5 & 31.5 & 35.6 \\
CLAN  &77.6 & 31.2 & 69.9 & 5.4 & 0.2 & 21.8 & 5.6 & 6.6 & 68.6 & - & 73.9 & 50.8 & 19.1 & 56.6  & - & 24.4 & - &  5.4 & 27.5 & 34.0 \\
CRST  & 62.2 & 23.4 & \textbf{75.6} & 6.4 & 1.6 & \textbf{33.8} & 22.8 & \textbf{30.9} & 77.8 & -  & 71.5 & 48.6 & 24.2 & 79.5 & - & 15.2 & - & 12.9 & 48.4 & 39.7 \\
CAG-UDA & 71.9 & 32.8 & 71.7 & 19.6 & 7.7 & 34.5 & 0.0 & 16.6 & 80.3 & - &  80.9 & 51.9 & \textbf{32.5} & 79.4 & - &  26.5 & - &  \textbf{22.1} & 51.9 & 42.5 \\
\rowcolor{lightgreen}
\textcolor{Orange}{DCAA}  & 67.7 & 30.7 & 74.7 & \textbf{22.9} & \textbf{14.1} & 33.2 & \textbf{38.5} & 30.4 & \textbf{82.4} & - & \textbf{81.6} & \textbf{60.1} & 31.2 & \textbf{80.8} &  - & \textbf{40.2} & - &  14.5 & \textbf{54.7} & \textbf{47.4} \\
\hline
\end{tabular}
\caption{\textbf{Performance on Cityscapes$^{*}$.} We reproduce other methods using their released codes following their best configurations. We clarify these reproduced scores under our setting are reliable considering the improvements over the baseline (Source Only). ++: We improve these CDA methods by providing the target sub-domain labels, and using our CGST.}
\label{tab:performance-cityscape}
\end{table*}

\subsubsection{Adversarial Ambivalence.} With the pseudo-labels, we can directly transfer knowledge in a more explicit manner, \emph{i.e.,} supervised learning:
\begin{equation}
    \begin{aligned}
      \tilde{\mathcal{L}}^{T}_{CE}(x^t) &= - \sum_{h,w}\sum_{l=1}^L y_{hwl}^t(x^t)\log(p^{CA}_{hwl}(x^t)).
    \end{aligned}
\end{equation}
However, the learned pseudo-labels are with many noises. Several previous works addressed the issue by adding regularization terms \cite{zou2019confidence} or discovering hard adaptation images \cite{pantwo2020}. For the latter, researchers compute the image-level confidence based on the pixel-level pseudo-label confidence, then employ pseudo-label supervision and adversarial learning for easy and hard samples, respectively. In contrast, we exploit the adversarial ambivalences by considering the dense discriminator outputs (PatchGAN) to further regulate the easy and hard adaptation regions. For each target image, we utilize pseudo-labels and the adversarial confidence to supervise the easy adaptation regions in a soft manner, and constraint the hard adaptation regions via adversarial loss. We take the regions where $y_{hw}^t (x^t) = 0$ as the hard adaptation regions.

Specially, we feed $p(x^t)$ into the discriminator $D^{CA}$ to obtain an adaptation confidence map $d(x^t)$. Then, we rectify $\tilde{\mathcal{L}}^{T}_{CE}(x^t)$ by incorporating pixel-wise weights as: 
\begin{equation}
    \begin{aligned}
      \mathcal{L}^{T}_{CE}(x^t) &= - \sum_{h,w}d_{hw}(x^t)\sum_{l=1}^L y_{hwl}^t(x^t)\log(p^{CA}_{hwl}(x^t)).
    \end{aligned}
\end{equation}
We filter out the easy adaptation regions to constraint the remained regions as:
\begin{equation}
    \begin{aligned}
      \mathcal{L}^{T}_{adv}(x^t) &= - \sum_{h,w}\delta(y_{hw}^t (x^t) = 0)\log(D^{CA}(p^{CA}(x_hw^t))).
    \end{aligned}
\end{equation}

In the self-training stage, $\mathcal{L}^{S}_{CE}$ and $\mathcal{L}^{T}_{CE}$ encourage the network produces domain invariant features, and $\mathcal{L}^{T}_{adv}$ allows the network focusing on pushing hard adaptation regions to the feature space. We set $\lambda_p$ to $0.6$ to generate more pseudo-labels for target images benefiting from the adversarial ambivalence mechanism. Besides, we adopt $M_0$ (with fixed parameters) to produce pseudo-labels and adversarial ambivalences. We learn a new model $M_1$ that shares the same network architecture with $M_0$. The parameters are initialized via the pre-trained $M_0$.  


\section{Experiments}

We refer to the supplemental materials for the implementation details and more analyses of our approach. Our codes will be made public available.

\begin{figure*}[th!]
\centering
\includegraphics[scale=0.27]{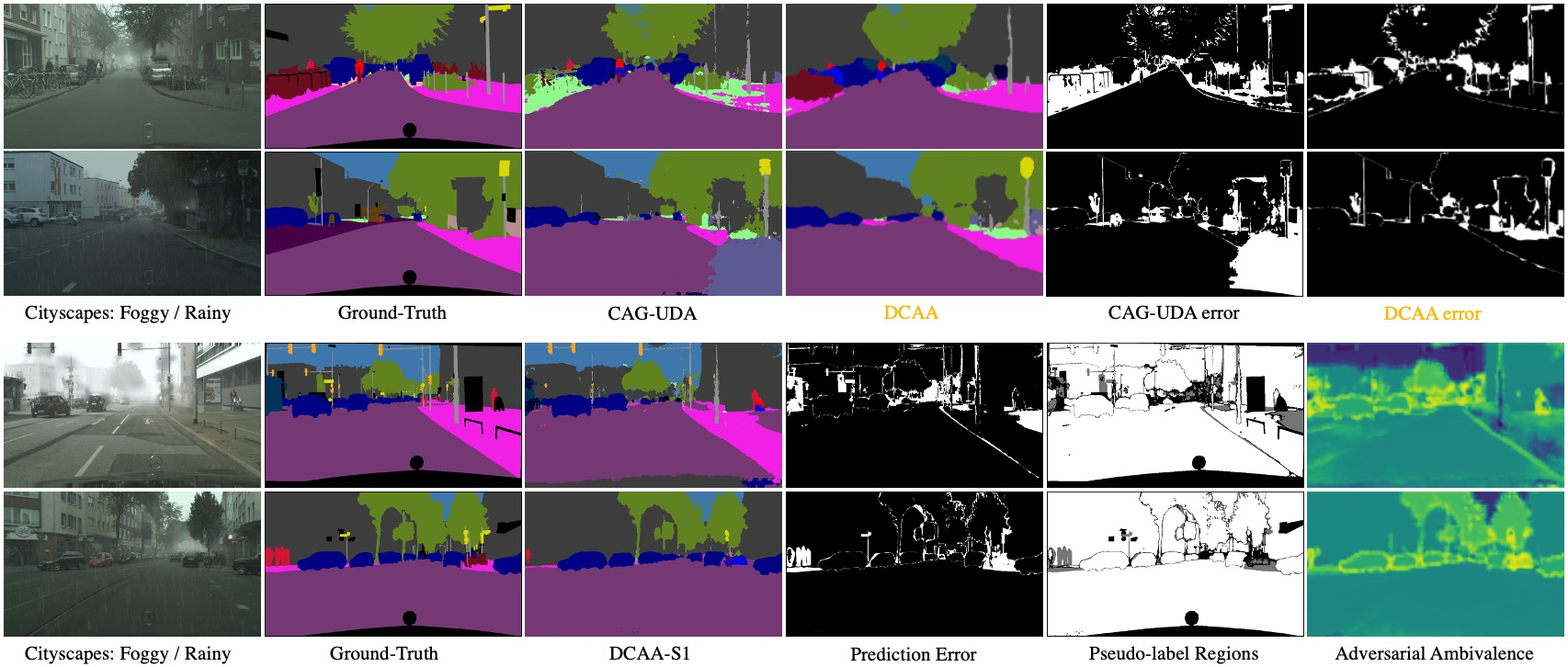}
\caption{\textbf{Qualitative Results.} We refer to Figure~\ref{fig:motivation} for the explanations of these maps. From the top part, our predictions contain more details compared with CAG-UDA. From the bottom part, the adversarial ambivalence is potent to indicate the incorrectly predicted regions and revise the pseudo-label related constraints.}
\label{fig:DCAA}
\end{figure*}

\subsubsection{Adaptation Scenarios.} We evaluate our proposed approach on three challenging adaptation scenarios, \emph{i.e.,} GTA5$\to$Cityscapes$^*$ and SYNTHIA$\to$Cityscapes$^*$, and GTA5$\to$BDD100K. Cityscapes$^*$ is a combined benchmark consisting of Cityscapes-cloudy (the commonly used Cityscapes dataset), Cityscapes-Foggy, and Cityscapes-Rainy. Specifically, Cityscapes-Cloudy provides 3,457 images with a resolution of $2048\times1024$, which are officially split into 2,975 training images and 500 validation images. The images in Cityscapes-Foggy are captured by simulating different levels of foggy to images in Cityscapes-Cloudy. Cityscapes-Rainy selects 262 training images and 33 validation images in Cityscapes-Cloudy to attach 36 degrees of rain to generate 9,432$\setminus$1,118 realistic rainy scenes. We randomly collect 12 rainy images for each reference image to construct our final Cityscpes-Rainy, \emph{i.e.}, 3,144 images for training, and 396 images for evaluation. The enlarged target benchmark (Cityscapes$^*$) is now has 9,094 ($2975 + 2975 + 3144$) images for training and 1,396 ($500 + 500 + 396$) images for evaluation. BDD100K is constructed for heterogeneous multitask learning where the images are with diverse weather conditions. We take BDD100K: $\{Rainy, Snowy, Cloudy\}$ as our target domain following OCDA \cite{liu2020open}. There are 4,855$\setminus$215, 5,307$\setminus$242, and 4,535$\setminus$346 training$\setminus$test images for the three weather conditions, respectively. We also evaluate our trained model on the unseen BDD100K-Overcast domain (627 images). The synthetic dataset GTA5 contains 24,966 images with a size of $1914\times1052$, and has the consistent 19 semantic categories as Cityscapes. SYNTHIA is another synthetic benchmark that offers 9,400 images of size $1914\times1052$. We use the 16 common category labels as Cityscapes for evaluation.

\setlength\tabcolsep{5pt}
\begin{table}[th!]
\footnotesize
\centering
\begin{tabular}{ c | c  c  c | c || c  }
\hline
Method & Rainy & Snowy & Cloudy & Avg. & Overcast \\
\hline
Source Only & 16.2 & 18.0 & 20.9 & 18.9 & 21.2 \\
AdaptSegNet & 20.2 & 21.2 & 23.8 & 22.1 & 25.1 \\ 
CBST & 21.3 & 20.6 & 23.9 & 22.2 & 24.7 \\
IBN-Net & 20.6 & 21.9 & 26.1 & 22.8 & 25.5 \\
PyCDA & 21.7 & 22.3 & 25.9 & 23.3 & 25.4 \\
OCDA & 22.0 & 22.9 & 27.0 & 24.5 & 27.9 \\
OCDA++ & 23.1 & 22.8 & 29.2 & 25.7 & 29.8 \\
MDTA-ITA & 22.6 & 23.2 & 27.5 & 24.9 & 24.9 \\
\rowcolor{lightgreen}
\textcolor{Orange}{DCAA} & 31.0 & 34.6 & 37.4 & \textbf{34.7} & 38.3 \\
\hline
\end{tabular}
\caption{\textbf{GTA5$\to$BDD100K.} During the training process, the target sub-domains consists of ``Rainy", ``Snowy", and ``Cloudy", while ``Overcast" is unseen. }
\label{tab:performance-bdd100k}
\end{table}

\subsubsection{Compared Methods.}
We reproduce several widely investigated SOTA methods, including AdaptSegNet \cite{tsai2018learning}, CLAN \cite{luo2019taking}, CRST \cite{zou2019confidence}, ADVENT \cite{vu2019advent}, and CAG-UDA \cite{zhang2019category} using their shared codes and their best configurations for benchmark performance reporting. Directly evaluate their methods using the public models will result in lower scores since their models are trained only for Cityscapes-Cloudy. Moreover, we make the sub-domain labels available to improve the open domain adaptation approaches
OCDA \cite{liu2020open} and AMEAN \cite{chen2019blending}. We also compare our DCAA with a SOTA multi-target domain adaptation method MDTA-ITA \cite{gholami2020unsupervised}.

\subsubsection{General Performance.} The scores are reported in Table~\ref{tab:performance-bdd100k} and Table~\ref{tab:performance-cityscape}. Generally, DCAA ranks $1st$ on all the studied adaptation scenarios. For GTA5$\to$BDD100K, DCAA achieves consistently promising improvements (about $9.0\%$) over all the target sub-domains, even for the unseen ``Overcast" images. The results demonstrate that our approach can effectively exploit multi-modalities' characteristics and is potent to produce high-performing adaptation models. For GTA5$\to$Cityscapes$^{*}$, DCAA outperforms the baseline (Source Only) by a large margin ($51.4\%$ vs. $30.6\%$), while yields a significant improvement ($6.0\%$) over the SOTA method CAG-UDA. Besides, we find that OCDA and MDTA-ITA can not perform well on the segmentation adaptation scenarios. For example, OCDA only captures slight improvements over the AdaptSegNet baseline ($1.1\%\sim2.4\%$) as reported in Table~\ref{tab:performance-cityscape} and Table~\ref{tab:performance-bdd100k}. A possible reason is that they focuses on disentangling the domain-style codes from the content codes. It may easy to disentangle global style codes for image classification problems as studied in their paper. However, extracting content independent pixel-level style codes would be more challenging. For SYNTHIA$\to$Cityscapes$^{*}$, DCAA produces appealing semantic parsing for target images with a remarkable mean IoU ($47.4\%$). Moreover, we present some qualitative results in Figure~\ref{fig:DCAA}. Our DCAA can preserve more details in boundary areas and revise some incorrectly labeled regions predicted by other methods, especially for images with bad weather conditions. Furthermore, we find that the mIoU reported in previous publications (training and evaluating on Cityscapes-Cloudy) are consistently higher (about $5\%$) than the reproduced versions (training and evaluating on Cityscapes$^{*}$). The large performance gap shows that trivially learning an adaptation model without explicitly exploiting the diverse weather conditions may not sufficiently discovering their correlations.

\setlength\tabcolsep{3pt}
\begin{table*}[h]
\footnotesize
\centering
\begin{tabular}{ c | c c c c c c c c c c c c c c c c c c c | c }
\hline
Method & \rotatebox{90}{Road} & \rotatebox{90}{SW} & \rotatebox{90}{Build} & \rotatebox{90}{Wall} & \rotatebox{90}{Fence} & \rotatebox{90}{Pole} & \rotatebox{90}{Light} & \rotatebox{90}{Sign} & \rotatebox{90}{Veg.} & \rotatebox{90}{Terrace} & \rotatebox{90}{Sky} & \rotatebox{90}{Person} & \rotatebox{90}{Rider} & \rotatebox{90}{Car} & \rotatebox{90}{Truck} & \rotatebox{90}{Bus} & \rotatebox{90}{Train} & \rotatebox{90}{Moter} & \rotatebox{90}{Bike} & \rotatebox{90}{\textcolor{red}{mIoU}} \\
\hline\hline
\multicolumn{21}{c}{Condition Attention Module (\textcolor{Aquamarine}{\textbf{Stage1 Score}})} \\
\hline
Mix-Head & 88.0 & 41.5 & 81.1 & 36.5 & 27.3 & 30.7 & 45.0 & 27.3 & 82.6 & 37.3 & 73.5 & 61.7 & 32.4 & 85.0 & 27.6 & 42.1 & 13.6 & 17.5 & 23.7 & 46.0 \\
Sep-Head & 83.6 & 41.1 & 76.9 & 31.8 & 29.4 & 30.5 & 44.2 & 30.8 & 83.1 & 28.6 & 77.4 & 60.7 & 32.4 & 84.9 & 25.0 & 40.5 & 11.0 & 17.9 & 30.2 & 45.3 \\
SM-Head & 85.1 & 41.3 & 79.0 & 33.7 & 30.2 & 31.2 & 45.3 & 30.3 & 82.7 & 30.1 & 75.8 & 61.2 & 31.9 & 84.8 & 25.1 & 39.1 & 9.6 & 19.8 & 29.5 & 45.6 \\
\rowcolor{lightgreen}
\textcolor{Orange}{CAM} & {89.9} & {46.4} & {82.5} & {37.7} & 29.3 & 34.1 & 45.7 & 25.4 & 83.1 & 33.5 & 76.8 & 62.3 & 32.9 & 85.3 & 28.3 & 39.9 & 2.6 & 23.3 & 37.4 & \textbf{47.2} \\
\hline\hline
\multicolumn{21}{c}{Condition-Specific Adversarial Training (\textcolor{Aquamarine}{\textbf{Building on CAM}})} \\
\hline
DAT-S1 & 84.0 & 38.0 & 80.1 & 28.2 & 29.8 & 28.9 & 44.9 & 31.7 & 82.7 & 28.9 & 77.4 & 61.2 & 30.7 & 82.2 & 22.2 & 30.4 & 17.3 & 22.8 & 39.8 & 45.3 \\
\textcolor{Orange}{CSAT-S1} & 89.9 & 46.4 & 82.5 & 37.7 & 29.3 & 34.1 & 45.7 & 25.4 & 83.1 & 33.5 & 76.8 & 62.3 & 32.9 & 85.3 & 28.3 & 39.9 & 2.6 & 23.3 & 37.4 & \textbf{47.2} \\
DAT-S2 & 88.3 & 42.0 & 83.7 & 37.3 & 26.8 & 27.4 & 43.1 & 34.8 & 85.5 & 33.1 & 79.2 & 60.8 & 30.1 & 85.2 & 29.9 & 41.8 & 14.7 & 23.3 & 41.6 & 47.8 \\
\rowcolor{lightgreen}
\textcolor{Orange}{CSAT-S2} & 90.5 & {52.8} & {84.9} & {38.7} & {30.4} & {37.6} & {45.5} & {42.4} & {84.5} & {45.5} & {80.7} & {63.8} & {36.2} & {86.3} & {34.9} & {45.1} & {19.7} & 14.9 & {42.5} & \textbf{51.4} \\
\hline\hline
\multicolumn{21}{c}{Attentive Pseudo-Label Assignment ($\lambda_p = 0.9, \tilde{\mathcal{L}}^{T}_{CE}$) (\textcolor{Aquamarine}{\textbf{Baseline = DCAA-S1}})} \\
\hline
Baseline & 89.9 & 46.4 & 82.5 & 37.7 & 29.3 & 34.1 & 45.7 & 25.4 & 83.1 & 33.5 & 76.8 & 62.3 & 32.9 & 85.3 & 28.3 & 39.9 & 2.6 & 23.3 & 37.4 & 47.2 \\
MaxV & 89.8 & 44.1 & 83.2 & 34.7 & 29.3 & 29.5 & 42.7 & 30.9 & 86.0 & 44.3 & 80.6 & 60.7 & 31.2 & 86.2 & 32.2 & 30.1 & 17.7 & 26.1 & 34.7 & 48.1 \\
MeanV & 89.6 & 48.4 & 83.6 & 38.8 & 30.1 & 27.1 & 38.9 & 28.4 & 86.1 & 43.0 & 79.8 & 62.4 & 29.4 & 85.8 & 37.0 & 32.7 & 16.7 & 32.7 & 34.9 & 48.7 \\
\rowcolor{lightgreen}
\textcolor{Orange}{APLA$^{+}$} &90.5 & 47.0 & 83.7 & 35.0 & 17.0 & 31.2 & 43.6 & 36.6 & 85.9 & 44.3 & 77.7 & 63.8 & 34.3 & 88.2 & 30.6 & 38.7 & 8.6 & 34.9 & 54.9 & \textbf{49.8} \\
\hline
\multicolumn{21}{c}{Adversarial Ambivalence + APLA ($\lambda_p = 0.6$)} \\
\hline
Baseline ($\tilde{\mathcal{L}}^{T}_{CE}$) & 88.2 & 44.8 & 82.5 & 36.0 & 22.5 & 34.1 & 43.0 & 36.8 & 84.5 & 44.9 & 77.5 & 62.0 & 25.7 & 86.2 & 29.5 & 37.6 & 1.1 & 29.6 & 50.7 & 48.3 \\
$\mathcal{L}^{T}_{CE}$  & 90.6 & 50.9 & 84.0 & 41.0 & 22.1 & 36.1 & 45.8 & 34.3 & 85.1 & 36.4 & 79.4 & 64.5 & 37.7 & 85.0 & 33.0 & 46.5 & 18.9 & 22.3 & 47.3 & 50.6 \\
\rowcolor{lightgreen}
$\mathcal{L}^{T}_{CE} + \mathcal{L}^{T}_{adv}$ &  90.5 & {52.8} & {84.9} & {38.7} & {30.4} & {37.6} & {45.5} & {42.4} & {84.5} & {45.5} & {80.7} & {63.8} & {36.2} & {86.3} & {34.9} & {45.1} & {19.7} & 14.9 & {42.5} & \textbf{51.4} \\
\hline
\end{tabular}
\caption{\textbf{Ablation Studies.} Si: The $i$th stage results. APLA$^{+}$: When studying APLA, we take $\lambda_p$ as $0.9$ to avoid introducing too many noisy pseudo-labels. MaxV and MeanV mean that we obtain the pseudo-labels via the simple max voting and mean voting operations.}
\label{tab:apat}
\end{table*}

\subsection{Ablation Studies}
We conduct ablation experiments based on the adaptation scenario GTA5$\to$Cityscapes$^{*}$ to discuss our DCAA. We refer to the supplemental materials for other studies.

\subsubsection{Condition Attention Module.} We examine some other possible options that explicitly incorporate the weather priors, including Mix-Head, Sep-Head, and SM-Head. The differences are shown in Figure~\ref{fig:cam}. From Table~\ref{tab:apat}, either SM-head ($45.6\%$) or Sep-Head ($45.3\%$) even slightly decreases the performance of Mix-Head by $0.4\% \sim 0.7\%$. This observation explicates that trivially incorporate weather priors may show negative influences. Instead, CAM ($47.2\%$) effectively exploits the special characteristics and correlations of multi-modalities.

\subsubsection{Condition-Specific Adversarial Training.} We study the effectiveness of the standard adversarial training strategy and our progressive condition-specific adversarial training policy in minimizing the source and target discrepancy. In Table~\ref{tab:apat}, DAT denotes that all the discriminators are standard domain classifiers to distinguish source features from target features. We find that the second-stage DAT-S2 ($47.8\%$) only reaches a slight better mIoU over our first-stage CSAT-S1 ($47.2\%$), and is marginally below CSAT-S2 ($51.4\%$). This observation demonstrates that directly aligning source and target feature distributions is challenging when the target domain is heterogeneous.

\subsubsection{Self-Training.} We make comparisons with different baselines and variants to discuss our self-training policy. The scores and compared methods are presented in Table~\ref{tab:apat}. We find that the produced pseudo-labels by MaxV and MeanV can improve the baseline by $0.7\%\sim1.3\%$, while our APLA$^+$ impressively yields a mIoU of $49.8\%$ ($+2.7\%$). This observation shows that trivial pseudo-label generation strategies may not sufficiently exploit the complementarities of multi-modal semantics. Furthermore, the revised $\mathcal{L}^{T}_{CE}$ outperforms $\tilde{\mathcal{L}}^{T}_{CE}$ by a convincible margin ($+2.3\%$) when the pseudo-labels are with many noises. At the same time, the proposed adversarial training for hard adaptation regions ($\mathcal{L}^{T}_{adv}$) can further boost the mIoU to $51.4\%$. The improvements demonstrate that the proposed adversarial ambivalence is potent to suppress the adverse impact of noisy pseudo-labels. It can also drive the network focusing on pushing hard adaptation regions to a shared feature space.


\section{Conclusion}
In the paper, we notice that images in a target domain may show diverse characteristics in typical domain adaptation studies. We clarify carefully reason about the multi-modalities may enhance the adaptation performance. Thus, we propose a novel adaptation approach DCAA that is empowered by an attentive progressive adversarial training mechanism and a self-training policy. The former migrate the source and target discrepancies in a progressive manner. The later first produces pseudo-labels for target images by investigating the correlations among target sub-domains, then exploiting the adversarial ambivalence to regularize the noisy pseudo-labels and constraint the hard adaptation regions. We conduct various comparisons and ablation studies to discuss our methods intensely. Experimental results on several challenging adaptation scenarios demonstrate the superiority of method in capturing high-performing models.

\bibliography{egbib}

\section{Implementation Details}
We train the style translator (CGST) and the task-specific network independently as previous style transfer based methods. The DCAA can be optimized in an end-to-end manner if we have GPU cards with large memories. For CGST, the network architectures of the generator and discriminators are the same as those in CycleGAN. The adopted parser is a light version UperNet50 with half number of convolutional channels. We add a convolutional layer with kernel size $1\times1$ on top of each discriminator as the target weather condition classifier. Besides, we apply the same learning strategy and hyperparameters as those in the shared CycleGAN codes, except that we train the translator for 30 epochs where the first 10 epochs with fixed learning rate, and the other 20 epochs with the linear decayed learning rate. We reduce all the images to the resolution of $1280\times640$, and train the model on a random crop of $480\times480$.

For the segmentation part, we perform the experiments (including the compared methods) based on the 16 stride DeeplabV3+ following CAG-UDA. In detail, we take ResNet-101 as the backbone (encoder), and the improved ASPP module as the decoder. We train our segmentation model (encoder and decoders) in two stages (Stage1: $\mathcal{L}_{adv} + \mathcal{L}_{CE}^{S}$, Stage2: $\mathcal{L}_{CE}^{S} + \mathcal{L}_{CE}^T + \mathcal{L}_{adv}^T$). The discriminators are optimized via standard classification losses as AdaptSegNet. For the first-stage training, the loss weight for $\mathcal{L}_{adv}$ is 0.001. The SGD solver is with the momentum and weight decay of 0.9 and 0.0005, respectively. The beginning learning rate is set to 0.0025, and gradually decreases during the training procedure following the polynomial decay strategy with a power of 0.9. We train the model for 20 epochs with batch size 4. For the second-stage training, we use the same learning strategies as the first-stage training procedure, except that we use the base learning rate of 0.002 and the batch size of 2. We also train the second-stage model for 20 epochs. The batch normalization parameters are fixed in this stage. The data augmentation methods are shared for both the two phases. Specially, we train our model on a random crop of $896 \times 512$ followed by a random horizontal flip operation. We train our models using two V100 GPU cards. We provide the core parts of our codes in the supplemental materials. 

\setlength\tabcolsep{2.8pt}
\begin{table*}[h]
\footnotesize
\centering
\begin{tabular}{ c | c c c c c c c c c c c c c c c c c c c | c }
\hline
Method & \rotatebox{90}{Road} & \rotatebox{90}{SW} & \rotatebox{90}{Build} & \rotatebox{90}{Wall} & \rotatebox{90}{Fence} & \rotatebox{90}{Pole} & \rotatebox{90}{Light} & \rotatebox{90}{Sign} & \rotatebox{90}{Veg.} & \rotatebox{90}{Terrace} & \rotatebox{90}{Sky} & \rotatebox{90}{Person} & \rotatebox{90}{Rider} & \rotatebox{90}{Car} & \rotatebox{90}{Truck} & \rotatebox{90}{Bus} & \rotatebox{90}{Train} & \rotatebox{90}{Moter} & \rotatebox{90}{Bike} & \rotatebox{90}{\textcolor{red}{mIoU}} \\
\hline\hline
\multicolumn{21}{c}{UniModal v.s. MultiModal Translation (\textcolor{Aquamarine}{\textbf{Mix-Head, Stage1 Score}})} \\
\hline
CycleGAN & 85.7 & 42.3 & 76.2 & 33.4 & 22.8 & 27.4 & 41.2 & 27.2 & 77.0 & 24.3 & 70.1 & 58.2 & 21.0 & 82.4 & 22.3 & 30.7 & 4.8 & 22.1 & 41.2 & 42.7 \\
MUNIT$^{\star}$  & 75.4 & 31.7 & 71.4 & 19.4 & 27.6 & 27.3 & 39.6 & 27.3 & {83.9} & 28.7 & {77.2} & 60.2 & 28.9 & 83.2 & {32.8} & 41.8 & 8.0 & 20.5 & 39.4 & 43.4 \\
\rowcolor{lightgreen}
\textcolor{Orange}{CGST$^{\star}$} & 86.8 & {42.6} & 80.1 & 35.2 & {27.7} & {32.7} & {45.1} & {28.9} & 82.9 & 34.8 & 71.1 & 61.3 & 28.0 & 82.0 & 25.5 & 36.1 & 9.5 & {24.2} & {38.5} & \textbf{46.0} \\
\textcolor{Orange}{CGST} & {88.0} & 41.5 & {81.1} & {36.5} & 27.3 & 30.7 & 45.0 & 27.3 & 82.6 & {37.3} & 73.5 & {61.7} & {32.4} & {85.0} & 27.6 & {42.1} & {13.6} & 17.5 & 23.7 & 46.0 \\
\hline\hline
\multicolumn{21}{c}{MultiModal Translation (\textcolor{Aquamarine}{\textbf{Sharing strategies in DCAA}})} \\ 
\hline
MUNIT-S1 & 76.5 & 27.1 & 76.6 & 18.1 & 31.3 & 28.0 & 40.6 & 29.5 & 83.0 & 31.4 & 79.3 & 59.6 & 29.9 & 82.0 & 26.7 & 39.0 & 7.4 & 19.9 & 38.6 & 43.3 \\
StarGAN-S1 & 77.8 & 38.4 & 73.5 & 24.0 & 28.5 & 29.5 & 41.7 & 29.5 & 81.1 & 24.2 & 79.0 & 58.8 & 30.7 & 83.6 & 21.2 & 38.0 & 22.7 & 20.0 & 34.6 & 44.0 \\
\textcolor{Orange}{DCAA-S1} & 89.9 & 46.4 & 82.5 & 37.7 & 29.3 & 34.1 & 45.7 & 25.4 & 83.1 & 33.5 & 76.8 & 62.3 & 32.9 & 85.3 & 28.3 & 39.9 & 2.6 & 23.3 & 37.4 & \textbf{47.2} \\
MUNIT-S2 & 84.2 & 41.0 & 81.6 & 23.2 & 24.4 & 28.0 & 40.2 & 28.5 & 85.0 & 36.8 & 82.9 & 59.8 & 28.6 & 83.4 & 24.5 & 50.5 & 3.5 & 21.2 & 38.4 & 45.6 \\
StarGAN-S2 & 88.2 & 36.5 & 83.1 & 31.3 & 23.4 & 27.6 & 42.0 & 30.7 & 84.6 & 31.3 & 84.3 & 58.9 & 33.5 & 84.8 & 31.9 & 52.3 & 32.6 & 22.1 & 38.2 & 48.3 \\
\rowcolor{lightgreen}
\textcolor{Orange}{DCAA-S2} & 90.5 & {52.8} & {84.9} & {38.7} & {30.4} & {37.6} & {45.5} & {42.4} & {84.5} & {45.5} & {80.7} & {63.8} & {36.2} & {86.3} & {34.9} & {45.1} & {19.7} & 14.9 & {42.5} & \textbf{51.4} \\
\hline\hline
\multicolumn{21}{c}{Adversarial Ambivalence - $\lambda_p$ (0.5 $\sim$ 0.9)} \\
\hline
Baseline ($\tilde{\mathcal{L}}^{T}_{CE}$)-0.5 & 88.4 & 42.6 & 82.8 & 36.5 & 19.8 & 34.5 & 41.8 & 32.5 & 86.2 & 46.5 & 81.2 & 63.7 & 26.1 & 86.6 & 21.3 & 20.7 & 4.3 & 29.6 & 52.2 & 47.2 \\
$\mathcal{L}^{T}_{CE}$-0.5  & 89.1 & 45.0 & 83.9 & 38.3 & 21.6 & 37.7 & 46.0 & 45.3 & 84.7 & 39.1 & 79.0 & 64.7 & 36.6 & 86.2 & 31.5 & 38.6 & 15.2 & 19.6 & 49.4 & 50.1 \\
\rowcolor{lightgreen}
$\mathcal{L}^{T}_{CE} + \mathcal{L}^{T}_{adv}$-0.5 &  90.0 & 49.1 & 84.0 & 40.1 & 28.9 & 38.1 & 47.5 & 43.2 & 85.5 & 38.4 & 78.4 & 63.6 & 36.5 & 86.3 & 31.6 & 48.2 & 7.8 & 18.3 & 45.9 & \textbf{50.6} \\
\hline
Baseline ($\tilde{\mathcal{L}}^{T}_{CE}$)-0.6 & 88.2 & 44.8 & 82.5 & 36.0 & 22.5 & 34.1 & 43.0 & 36.8 & 84.5 & 44.9 & 77.5 & 62.0 & 25.7 & 86.2 & 29.5 & 37.6 & 1.1 & 29.6 & 50.7 & 48.3 \\
$\mathcal{L}^{T}_{CE}$-0.6  & 90.6 & 50.9 & 84.0 & 41.0 & 22.1 & 36.1 & 45.8 & 34.3 & 85.1 & 36.4 & 79.4 & 64.5 & 37.7 & 85.0 & 33.0 & 46.5 & 18.9 & 22.3 & 47.3 & 50.6 \\
\rowcolor{lightgreen}
$\mathcal{L}^{T}_{CE} + \mathcal{L}^{T}_{adv}$-0.6 &  90.5 & {52.8} & {84.9} & {38.7} & {30.4} & {37.6} & {45.5} & {42.4} & {84.5} & {45.5} & {80.7} & {63.8} & {36.2} & {86.3} & {34.9} & {45.1} & {19.7} & 14.9 & {42.5} & \textbf{51.4} \\
\hline
Baseline ($\tilde{\mathcal{L}}^{T}_{CE}$)-0.7 & 89.8 & 47.0 & 83.5 & 38.6 & 24.5 & 35.5 & 44.7 & 35.1 & 84.2 & 34.3 & 75.7 & 64.0 & 26.9 & 87.4 & 25.8 & 35.3 & 3.4 & 21.6 & 52.7 & 47.9 \\
$\mathcal{L}^{T}_{CE}$-0.7  & 90.5 & 48.8 & 84.7 & 38.0 & 31.3 & 38.8 & 47.0 & 41.2 & 84.7 & 34.9 & 80.6 & 64.3 & 37.2 & 85.8 & 31.5 & 40.8 & 16.7 & 20.4 & 47.9 & 50.8 \\
\rowcolor{lightgreen}
$\mathcal{L}^{T}_{CE} + \mathcal{L}^{T}_{adv}$-0.7 &  90.9 & 50.1 & 83.8 & 39.2 & 23.2 & 36.5 & 46.3 & 42.3 & 85.6 & 43.8 & 78.1 & 64.6 & 37.5 & 86.3 & 29.5 & 40.7 & 8.8 & 30.2 & 52.6 & \textbf{51.1} \\
\hline
Baseline ($\tilde{\mathcal{L}}^{T}_{CE}$)-0.8 & 88.5 & 46.4 & 83.2 & 37.5 & 24.2 & 37.5 & 45.1 & 39.3 & 83.4 & 37.7 & 79.7 & 64.5 & 36.5 & 86.0 & 27.2 & 34.4 & 4.2 & 20.2 & 48.2 & 48.6 \\
$\mathcal{L}^{T}_{CE}$-0.8  & 89.9 & 48.9 & 84.6 & 40.0 & 23.2 & 37.8 & 46.3 & 40.7 & 86.2 & 42.0 & 79.7 & 63.3 & 36.6 & 85.7 & 29.6 & 40.8 & 10.4 & 20.3 & 45.6 & 50.1 \\
\rowcolor{lightgreen}
$\mathcal{L}^{T}_{CE} + \mathcal{L}^{T}_{adv}$-0.8 &  91.4 & 50.7 & 84.1 & 38.6 & 23.4 & 36.5 & 45.8 & 41.1 & 85.8 & 45.1 & 77.4 & 64.8 & 36.5 & 86.2 & 28.7 & 44.8 & 3.0 & 28.2 & 51.2 & \textbf{50.7} \\
\hline
Baseline ($\tilde{\mathcal{L}}^{T}_{CE}$)-0.9 & 90.5 & 47.0 & 83.7 & 35.0 & 17.0 & 31.2 & 43.6 & 36.6 & 85.9 & 44.3 & 77.7 & 63.8 & 34.3 & 88.2 & 30.6 & 38.7 & 8.6 & 34.9 & 54.9 & 49.8 \\
$\mathcal{L}^{T}_{CE}$-0.9  & 89.8 & 48.9 & 83.5 & 38.7 & 21.2 & 37.3 & 44.5 & 36.3 & 85.5 & 42.6 & 79.2 & 65.0 & 37.4 & 85.6 & 28.0 & 36.5 & 10.9 & 22.9 & 48.2 & 49.6 \\
\rowcolor{lightgreen}
$\mathcal{L}^{T}_{CE} + \mathcal{L}^{T}_{adv}$-0.9 &  91.2 & 49.8 & 84.4 & 40.3 & 25.0 & 35.7 & 45.3 & 39.7 & 86.3 & 46.3 & 77.2 & 64.8 & 34.8 & 86.1 & 26.5 & 38.5 & 2.9 & 28.5 & 50.8 & \textbf{50.2} \\
\hline\hline
\multicolumn{21}{c}{Limitation: Teacher-Student Mechanism} \\ 
\hline
DCAA (Teacher) & 90.5 & {52.8} & {84.9} & {38.7} & {30.4} & {37.6} & {45.5} & {42.4} & {84.5} & {45.5} & {80.7} & {63.8} & {36.2} & {86.3} & {34.9} & {45.1} & {19.7} & 14.9 & {42.5} & {51.4} \\
DCAA$^{\dag}$ (Student) & 90.6 & 52.6 & 84.9 & 39.2 & 24.4 & 37.9 & 46.4 & 41.9 & 85.2 & 45.4 & 79.2 & 65.6 & 37.9 & 86.0 & 34.1 & 43.4 & 15.6 & 15.4 & 44.4 & 51.1 \\
\hline
\end{tabular}
\caption{\textbf{More Ablation Studies.} $\star$: For a fair comparison, we sample identical number of translated images (24,966) as CycleGAN to perform MUNIT$^{\star}$ and CGST$^{\star}$. Si: The $i$th stage results. Lower $\lambda_p$ means that the obtained pseudo labels will have more noises.}
\label{tab:ablation-supp}
\end{table*}

\section{More Ablation Studies}
\subsubsection{Condition Guided Style Transfer.} 
\begin{figure*}[th!]
\centering
\includegraphics[scale=0.55]{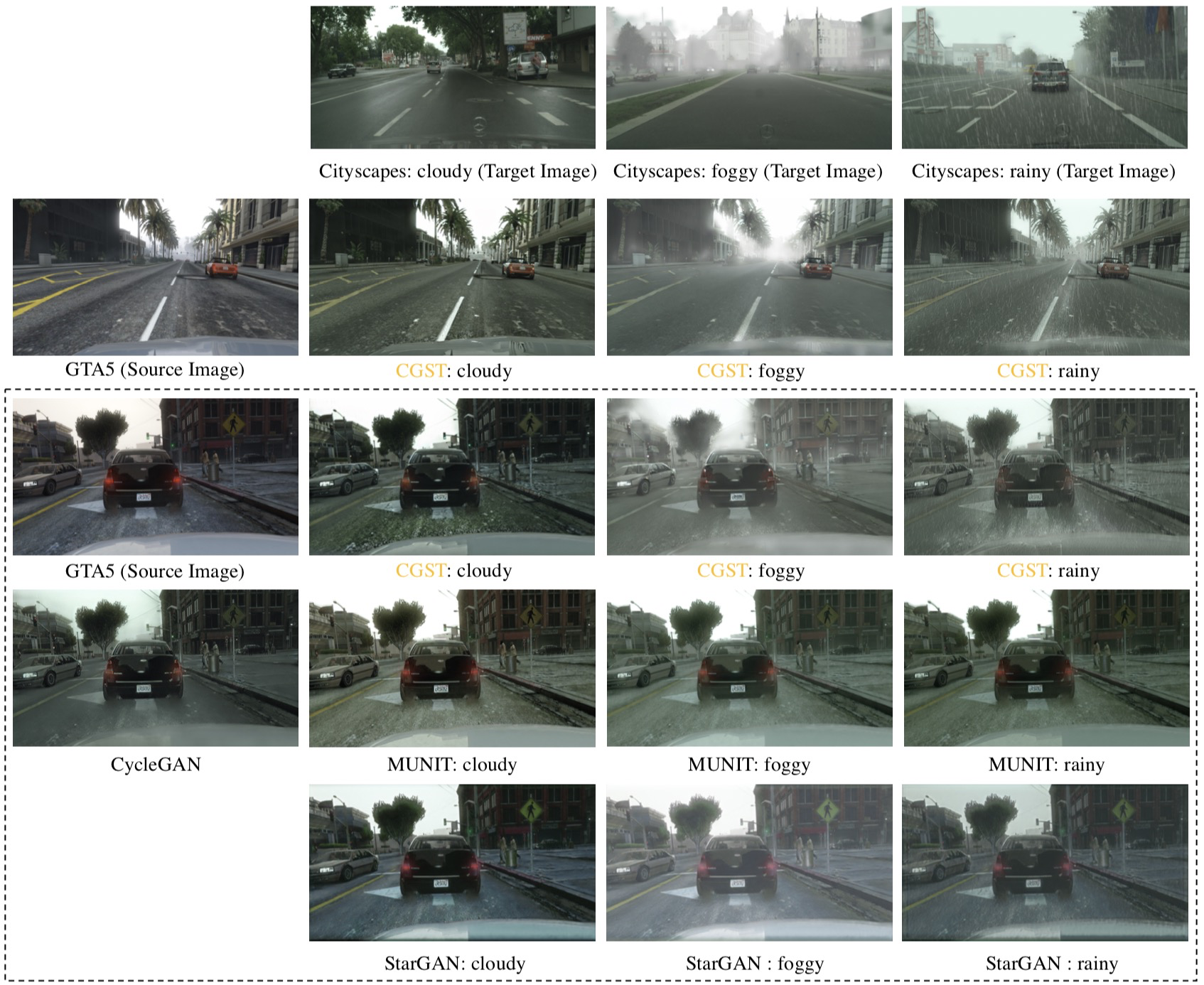}
\caption{\textbf{Condition Guided Style Transfer.} Fitst row: Example images (Cityscapes) under different weather conditions.}
\label{fig:translation}
\end{figure*}

We analyze the outcomes of UniModal and MultiModal translation networks in exploiting the weather characteristics for robust UDA. The scores are presented in Table~\ref{tab:ablation-supp}. We take the well studied CycleGAN \cite{zhu2017unpaired} and MUNIT \cite{huang2018multimodal} as the unimodal and multimodal framework examples, respectively. We incorporate $\mathcal{L}_{sc}$ into all the translation networks to prevent semantic distortions.  For MUNIT, we translate GTA5 to GTA5-Trans.: $\{Cloudy, Rainy, Foggy\}$ via the learned style translator. Taking GTA-Trans.:$\{Rainy\}$ as an example, for each GTA5 image, we randomly choose a rainy image from Cityscapes$^{*}$ to extract the style code, and using it to guide the translation. Our CGST$^{\star}$ ($46.0\%$) secures a promising improvement ($+3.3\%$) on mIoU compared with CycleGAN ($42.7\%$). We notice that MUNIT$^{\star}$ only performs slightly better ($+0.7\%$) than CycleGAN. This is because MUNIT has not encoded the weather codes well. We present some qualitative trainslations in Figure~\ref{fig:translation}. We find that CGST can offer visually more preferable results with natural styles. However, MUNIT and CycleGAN prefer to generate cloudy images. They show some limitations in handling bad weather. A possible reason is that, without a strong weather-related constraint, the adversarial learning process may focus more on distinguishing the style characteristic while overlooking weather characteristics. Besides, MUNIT-S1 and CGST capture the similar mIoU as MUNIT$^\star$ and CGST$^\star$. This observation shows that the enlarged source-domain training set may not bring more performance gains. Moreover, we make a comparison with the MultiModal translation baselines. Our DCAA ($51.4\%$) yields remarkably higher mIoU ($+3.1\% \sim 5.8\%$) than MUNIT ($45.6\%$) and StarGAN \cite{choi2018stargan} ($45.6\%$).

\subsubsection{Adversarial Ambivalence - $\lambda_p$} We study the impact of different $\lambda_p$ (Eqn. 9) for our self-training policy. The scores are reported in Table~\ref{tab:ablation-supp}. $\lambda_p$ is a trade-off parameters for the density and reliability of the pseudo labels. For most cases, we find that the proposed adversarial ambivalence is potent to suppress the adverse impact of noisy pseudo-labels ($\mathcal{L}^{T}_{CE}$ v.s. $\tilde{\mathcal{L}}^{T}_{CE}$). It can also drive the network focusing on pushing hard adaptation regions to a shared feature space ($\mathcal{L}^{T}_{adv}$). However, when we set $\lambda_p$ to 0.9, the revised $\mathcal{L}^{T}_{CE}$ yields a slightly lower mIoU than $\tilde{\mathcal{L}}^{T}_{CE}$. A possible reason is that most of the captured high-confidence pseudo labels are reliable. Thus, rectifying the pseudo labels via the adversarial ambivalences is unnecessary.

\section{Limitations}
One of the main limitations of our DCAA is that we need to learn multiple segmentation heads for each weather condition in a single framework. This may slightly degrade the inference efficiency. Luckily, we can partially address the issue by introducing a simple but effective teacher-student mechanism. In detail, we take the trained DCAA ($\lambda_p = 0.6$) as the teacher network with fixed parameters, and further learn a Mix-Head as the student network where the encoder is initialized via the pre-trained DCAA. The student network is optimized by aforementioned $\mathcal{L}^{S}_{CE}$ and $\mathcal{\tilde{L}}^{T}_{CE}$. We take the pseudo-labels captured by APLA of the teacher network as labels for target images ($\lambda_p = 0.9$). The simple teacher-student mechanism helps us capture a more efficient model DCAA$^{\dag}$ with a slight performance decrement (-0.3$\%$) as reported in Table~\ref{tab:ablation-supp}.

\section{More Qualitative Results}
\begin{figure*}[th!]
\centering
\includegraphics[scale=0.47]{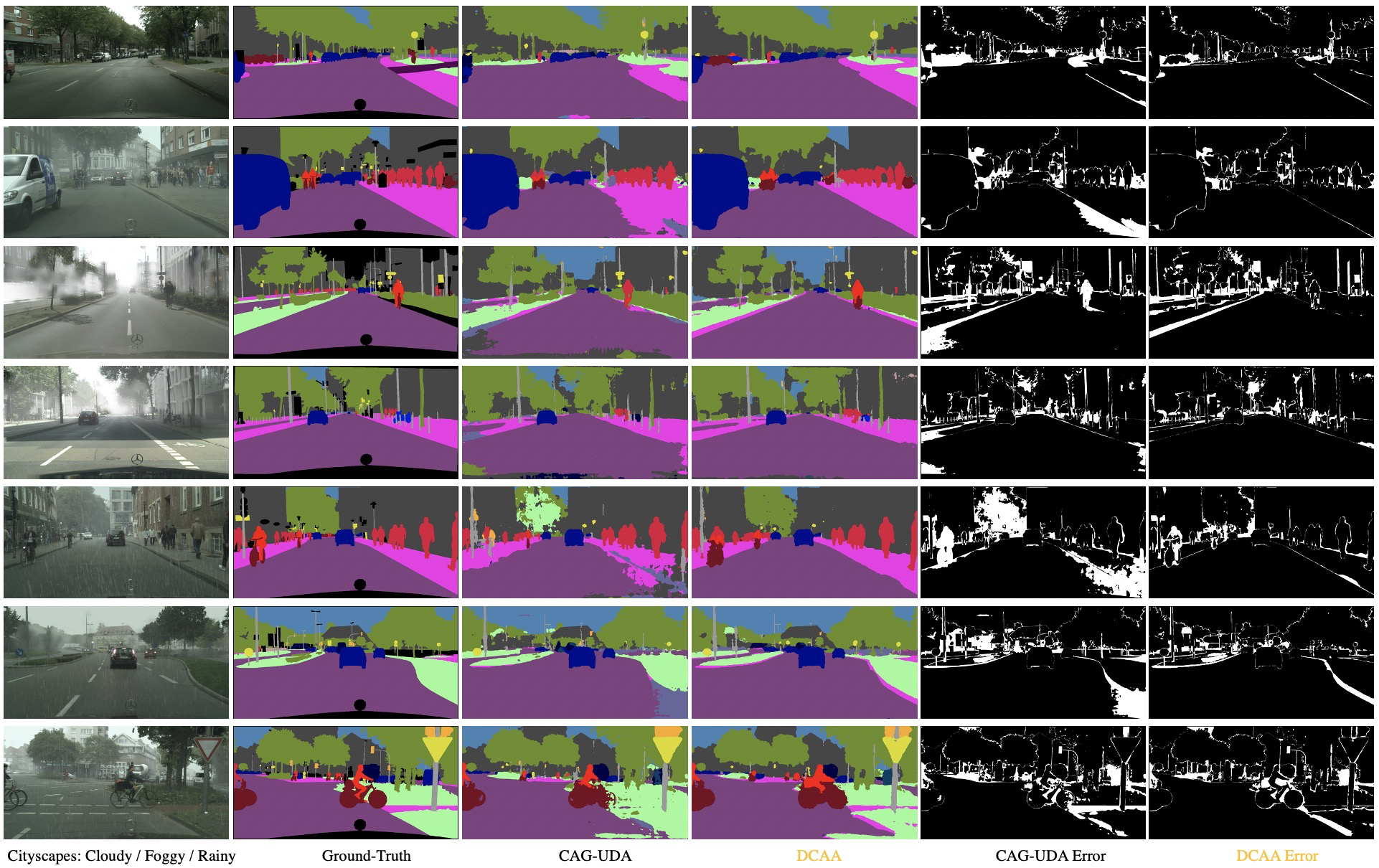}
\caption{\textbf{GTA5$\to$Cityscapes$^*$.} Zoom in for better view.}
\label{fig:cityscape}
\end{figure*}

\begin{figure*}[th!]
\centering
\includegraphics[scale=0.46]{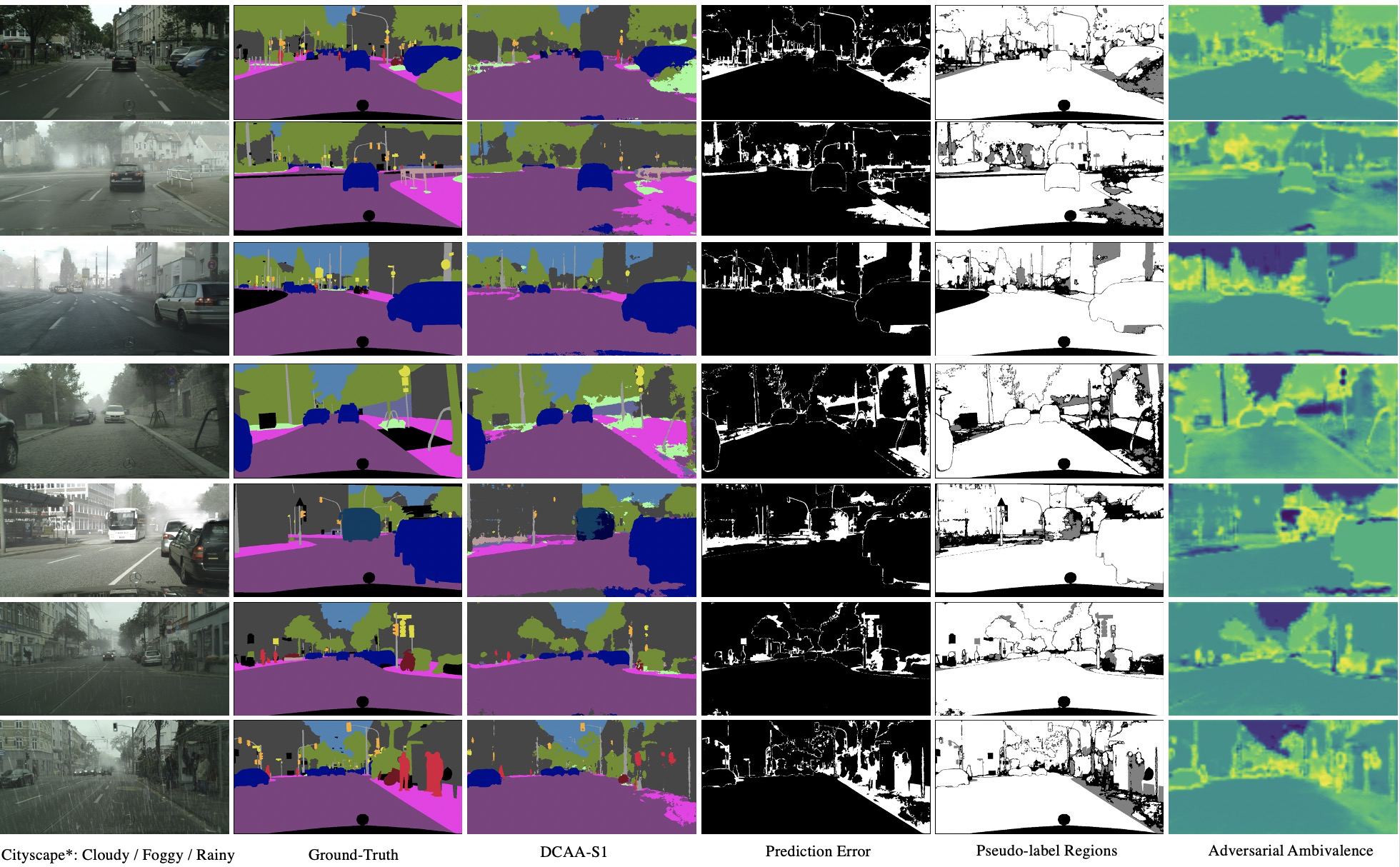}
\caption{\textbf{GTA5$\to$Cityscapes$^*$.} Zoom in for better view.}
\label{fig:cityscape-error}
\end{figure*}

\begin{figure*}[th!]
\centering
\includegraphics[scale=0.46]{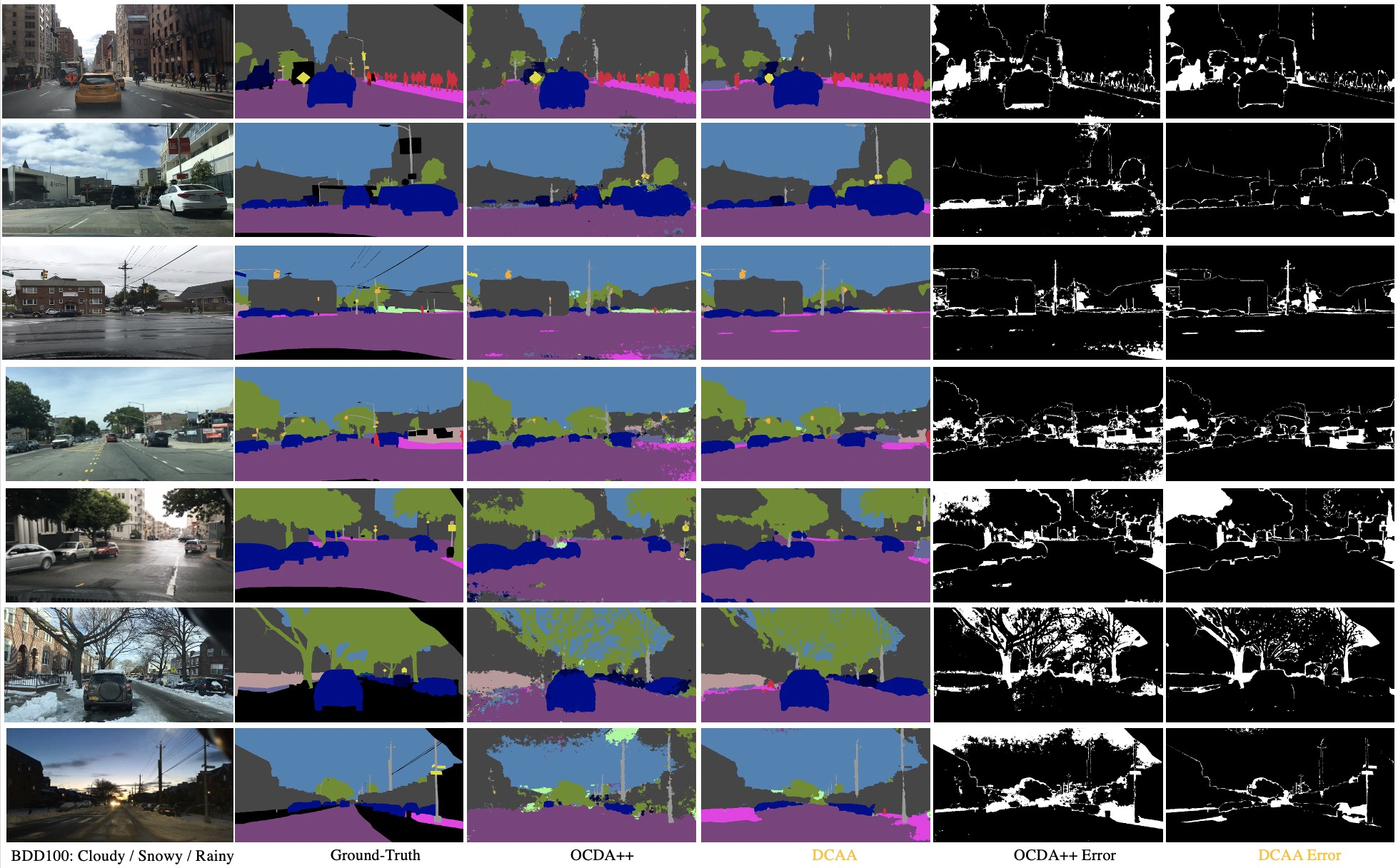}
\caption{\textbf{GTA5$\to$BDD100K.} Zoom in for better view.}
\label{fig:bdd100k}
\end{figure*}

\begin{figure*}[th!]
\centering
\includegraphics[scale=0.46]{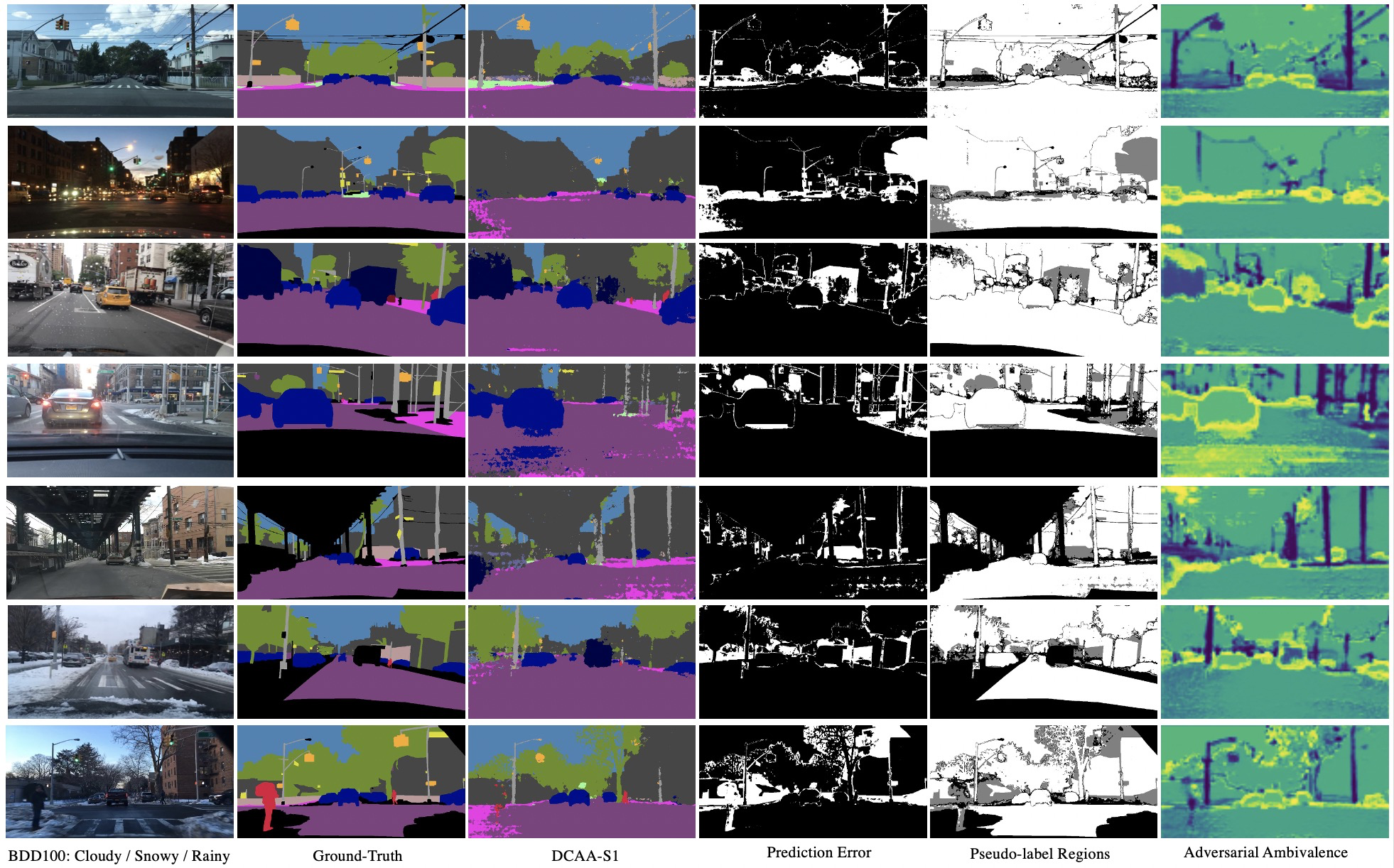}
\caption{\textbf{GTA5$\to$BDD100K.} Zoom in for better view.}
\label{fig:bdd100k-error}
\end{figure*}

We present more qualitative results in Figure~\ref{fig:cityscape} and Figure~\ref{fig:bdd100k}.  Our DCAA can preserve more details in boundary areas and revise some incorrectly labeled regions predicted by other methods, especially for images with bad weather conditions. We also show the consistency between the adversarial ambivalence and the prediction error in Figure~\ref{fig:cityscape-error} and Figure~\ref{fig:bdd100k-error}. We refer to our submission for more details.

\end{document}